# ROBOTICS PROJECT B31XP

Detecting and avoiding frontal obstacles from monocular camera for micro unmanned aerial vehicles

WakaWaka Group

Supervisor: Yvan Petillot

Team Members: H.Kidane, I.Sadek, M.Elawady

Heriot Watt University

School of Electrical and Physical Sciences

# 1 Contents

# 2 Introduction

Unmanned Aerial Vehicles (UAVs) or drones as some call them have been used in a large number of applications such as reconnaissance, surveillance, inspection, exploration, search and rescue. Despite a considerable amount of attempts have been carried out in order to make the process totally autonomous, obstacle avoidance is still too hard to deal with. UAVs can only carry very light weight and small size sensors like a monocular sensor for the reason of obstacle detection and avoidance. On the other hand, it is impractical to allow laser range finder or Kinect Microsoft cameras because they are all heavy which will increase power consumption and decrease time of flight [1].

There exists a lot of work in the literature trying to make the UAVs fly autonomously for example extracting perspective cues such as straight lines. However, it is only available in well-defined human made environments, in addition to many other cues which require enough texture information. Our main target is to detect and avoid frontal obstacles from a monocular camera using a quad rotor Ar.Drone 2 by exploiting optical flow as a motion parallax, the drone is permitted to fly at a speed of 1 m/s and an altitude ranging from 1 to 4 meters above the ground level as shown in fig.1. In general, detecting and avoiding frontal obstacle is a quite challenging problem because optical flow has some limitation which should be taken into account i.e. lighting conditions and aperture problem.

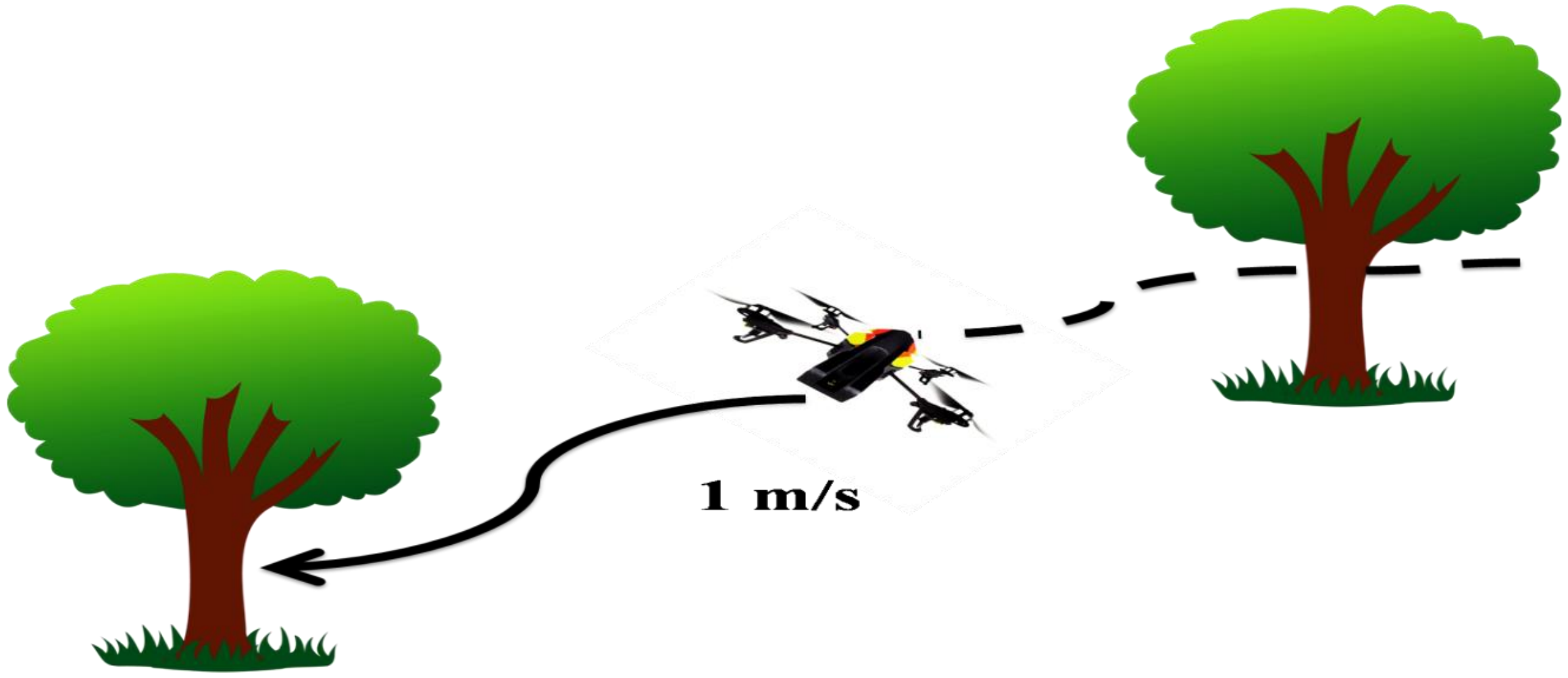

Fig.1 A toy example shows Ar.Drone flying in an outdoor environment at a constant speed 1 m/s

# 3 Description of UAV Platform

Ar.Drone 2 is a rotating rigid structure with six degree of freedom. The mechanical structure consists of four rotors. Each pair of opposite rotors rotates in the same way. One pair is rotating clockwise and the other is rotating counter-clock wise. The main parts can be shown in Fig.2:

- HD camera 720 pixels records video at 30 frames per second, support H.264 recording format and JPEG photo capture
- Wide angle lens 92° diagonal, the focal length is smaller than normal lenses to include more information about the scene
- Ultrasound sensor which provides ground altitude measurement up to six meters above the ground

- Very light and high resistance plastic that protect the drone in indoor environments

For more technical information reader can refer to the officialAr.Drone2 website.

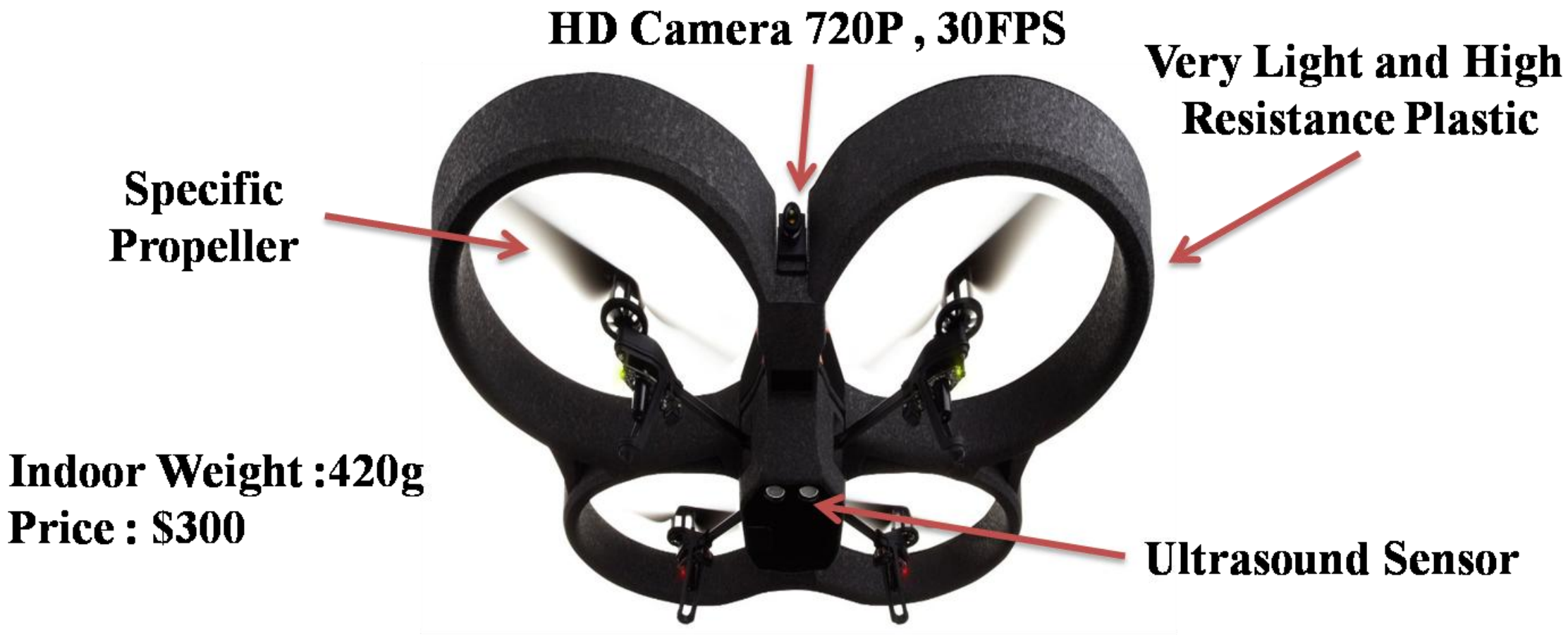

Fig.2 Ar.Drone2 main parts

# 4 Related work

Plentiful amount of work have been done to autonomously navigate UAVs in unrecognized environments. Three mechanisms can be utilized by human eyes for3D depth estimation: moving parallax, monocular cues, and stereo vision.

Moving parallax is exploited by optic flow in several aspects such as [2] developed an autonomous UAV navigation by exploiting optical flow along with neural network to adjust controller gain in indoor environments. The quality of optical flow can be altered by many reasons i.e. light intensity, contrast and texture which are the three most important factors. Thanks to neural network such kind of data can be dealt with by learning a neural network. In order to solve the problem of automatic control gain, a multilayer feed-forward back-propagation neural network has been utilized. It is learned and tested via two inputs. Firstly, one dimensional optical flow of the model provided by optical flow sensor. Secondly, the intensity of surrounding lighting condition by a digital light sensor. Changing the florescent lighting condition affords various phases for training. The output of optical flow is represented in terms of distance to the obstacle, by introducing this idea time consuming calibration procedures can be avoided.

Strategy to bypass frontal and lateral obstacles by using optical flow (correlation based) is presented by [3] for the sake of a rotary wing (realistic simulated helicopter) control in 3D urban texture environment. Two control levels have been used, Low level controller to adjust a straight line flight and to keep a constant speed and high-level obstacle-avoidance controller which determines the optical flow component required to avoid obstacles. Once again lighting condition should be taken into account since it contributes to obstacle collision behavior.

In the same manner of Muratet et al, [4] suggested autonomous micro-flyers which mimic fly behavior at three different levels. At the sensory level, a low resolution sensor is used as the experimental environment is already equipped by high contrast texture on the wall and micro-electro-mechanical system (MEMS) rate gyro that can provide control signal during rotation. The contribution of this paper appears in the signal processing level where optical flow estimation is fused with inertial information provided by inertial sensors to maintain robot balance functioning as flies (cancel the rotational optical flow). At the behavioral level, the designed control system gives straight movements combined with rapid turning in case of fast approaching collision. Similar ideas can be found in [5], [6].

A method to compute structure from motion that doesn't need an accurate calculation of optical flow of each feature point is introduced by [7] to precisely compute an individual optical flow value for each feature point is not easy to achieve this is owing to the image noise effect and the aperture problem. Alternatively they proposed the idea of optical flow probability distribution. The proposed method can be appropriate in case of points with sharp edges. Structure from motion is calculated using two different approaches, linear structure from motion, to overcome nonlinear optimization problem and optimal structure from motion, which is designed to handle the existence of noise. To sum up all, the covariance estimate (how accurate the optical flow detection is in any direction) should be a necessary step for any optical flow technique.

[1] An iterative training procedure called data set aggregation (DAgger) is used. This method trains a policy which observes a human expert pilot an Air Drone. They extract visual features i.e. (radon like features, structure tensor, Laws' masks and Optic flow) which serve as input to the unsupervised learner. After some iteration the Dagger can assure a policy that mimics a human expert command.

Monocular cues, human eyes have the ability to judge depths by using only a single image for example, [8] come up with an idea to navigate UAV in indoor environments just as home and office building. First, the classifier identifies the type of the environment. In case of a corridor the aim is to detect edges by canny edge detector, and then to find long lines using probabilistic Hough transform because at the end of the corridor long lines can been seen as a vanishing point. For the stair case , the purpose is to find center of the stair case by utilizing same steps such as in the corridor then classify line segments into horizontal and vertical lines. Compute the mean of end points only for the largest horizontal line cluster which can be considered as an indicator for the vehicle to fly. This scheme is best suits to well defined structured environments where as it will fails in outdoor environments due to the absence of straight line.

[9] implemented an algorithm on the basis of relative size change i.e. assume two objects have common sizes while the robot is moving capturing scene information if the size of one object is increasing it means that this object is close to the moving robot. The principle idea is to extract SURF features from consecutive frames, and then match similar features, discard false match in terms of Euclidean distance threshold also discard features which have the same or similar sizes. Finally, confirm scale with template matching to compute the position of the obstacle.  This method works fine if the environment is rich in texture information.

[10] Many other methods depending on monocular cue can be found in literature for instance, texture gradient: the closer the object the more texture information we can notice on its surface, occlusion: objects which partially or fully block the appearance of other objects are sensed as closer in depth, haze: because of the atmospheric light scattering the farther the object the hazier it look so distal objects can be perceived as blurred, shadow also plays an important role to determine the shape of the object and its size in space. These approaches require good features for defined objects.

Stereovision is to extract 3D information from digital images by means of triangulation and use this information to estimate position of objects in two images. Stereovision requires sufficient base line separation between the lenses as a small error in triangulation results in a large error in distance calculation [11].

# 5 Methodology

First of all, a brief introduction is given about optic flow. The optical flow can be defined as the apparent motion between two consecutive frames [12].In the time of flying, objects which are close to the drone have higher optical flow magnitudes. [Oh et al., 2004]  If we consider flying insects, they avoid immediate collision with what is called saccade or turning away from high optical flow regions as presented in fig.3.

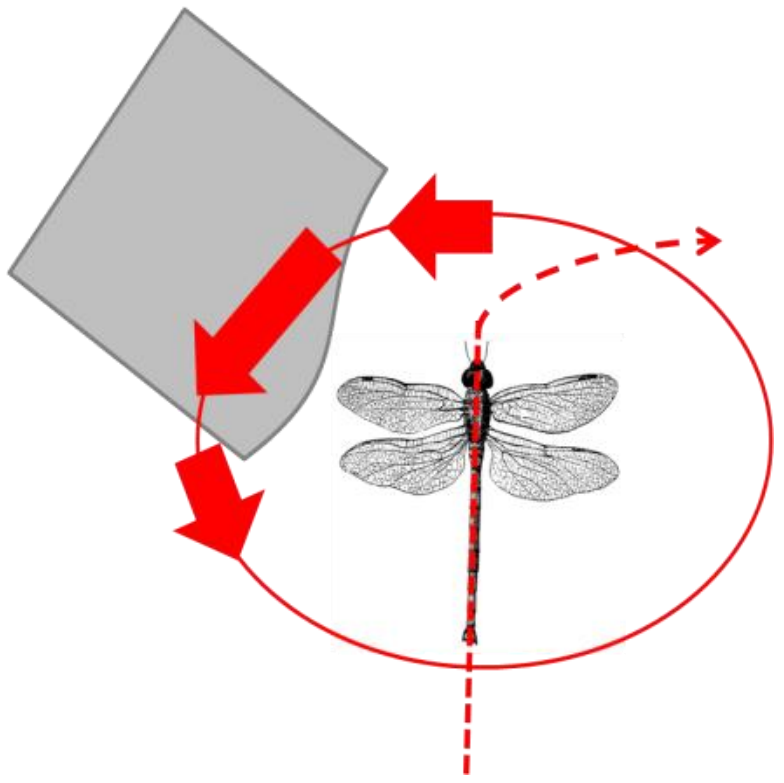

Fig.3 Flying insect (dragonfly) avoid an obstacle by turning away from high optical flow region [Oh et al., 2004]

Our implementation is based on computing a dense optical flow by applying Farneback's algorithm [13]. The estimation of displacement is carried out in different steps which we will try to summarize based on the original paper as follows:

- Polynomial estimation

The point is to approximate a neighboring of each pixel by means of polynomial. This step is only applied spatially not with the time.

$$f(x) \sim x^T A x + b^T + c \qquad \textbf{Eq.1}$$

Where $A$ is defined as a symmetric matrix and $b$ a victor and $c$ a scalar value.

- Displacement estimation

The purpose is to explorer the effect of displacement on the polynomial function.

$$f_1(x) \sim x^T A_1 x + b_1^T + c_1 \qquad \textbf{Eq. 2}$$

Create another signal by considering a global displacement $d$

$$\begin{aligned} f_2(x) = f_1(x-d) &= (x-d)^T A_1 (x-d) + b_1^T (x-d) + c_1 \\ &= x^T A_1 x + (b_{1-} A_1 d)^T x + d^T A_1 d - b_1^T d + c_1 \qquad \textbf{Eq.3} \\ &= x^T A_2 x + b_2^T + c_2 \end{aligned}$$

Comparing coefficients in the quadratic polynomials gives

If $A_1$ is assumed to be a non singular matrix, the value of the translation $d$ can be calculated,

$$A_2 = A_1 \qquad \textbf{Eq.4}$$

$$b_2 = b_1 - 2A_1 d \qquad \textbf{Eq. 5}$$

$$c_2 = d^t A_1 d - b_1^T d + c_1 \qquad \textbf{Eq.6}$$

$$2A_1 d = -(b_2 - b_1) \qquad \textbf{Eq.7}$$

$$d = -\frac{1}{2} A_1^{-1} (b_2 - b_1) \qquad \textbf{Eq.8}$$

According to the paper this value holds for any dimension.

- Practical consideration

The belief that the whole signal is a polynomial and a global translation describing the two signals are impractical. The solution is to replace the global assumption in equation 2 with a local one.

$$A(x) = \frac{A_1(x) + A_2(x)}{2} \qquad \textbf{Eq. 9}$$

$$\triangle\, b(x) = -\tfrac{1}{2}(b_1(x) - b_2(x)) \qquad \textbf{Eq. 10}$$

$$\mathrm{A(x)d(x)} = \triangle\, b(x) \qquad \textbf{Eq. 11}$$

Where

$A_1(x)$ , $b_1(x)$ *and* $c_1(x)$ , are expansion coefficients for first image

$A_2(x)$ , $b_2(x)$ *and* $c_2(x)$ , are expansion coefficients for first image

$d(x)$Implies that the global displacement is substituted by local spatial displacement

- Estimation over a Neighborhood

Equation 11 can be solved by minimizing below function:

$$\sum_{\Delta x \in I} w(\Delta x) \parallel A(x+\Delta x)d(x) - \Delta b(x+\Delta x) \parallel^2 \quad \textbf{Eq.12}$$

$w(\Delta x)$ , is a weight function for neighboring points. The minimums is computed for

$$d(x) = (\sum wA^T A)^{-1} \sum wA^T \Delta b \quad \textbf{Eq.13}$$

$$e(x) = (\sum w \Delta b^T \Delta b) - d(x)^T \sum wA^T \Delta b \quad \textbf{Eq.14}$$

$e(x)$ , is the minimum value and the coefficients in equation 14 are computed by point wise operation. This method can be applied in two different ways.

| | Problem | Solution |
|---|---|---|
| **Iterative** | Too large displacement means the output will not be improved and iterating will be senseless | Coarse scale analysis i.e. low pass the signal in the first place |
| **Multi-scale** | For each scale , the polynomial coefficients should be computed | Sub sampling between scales |

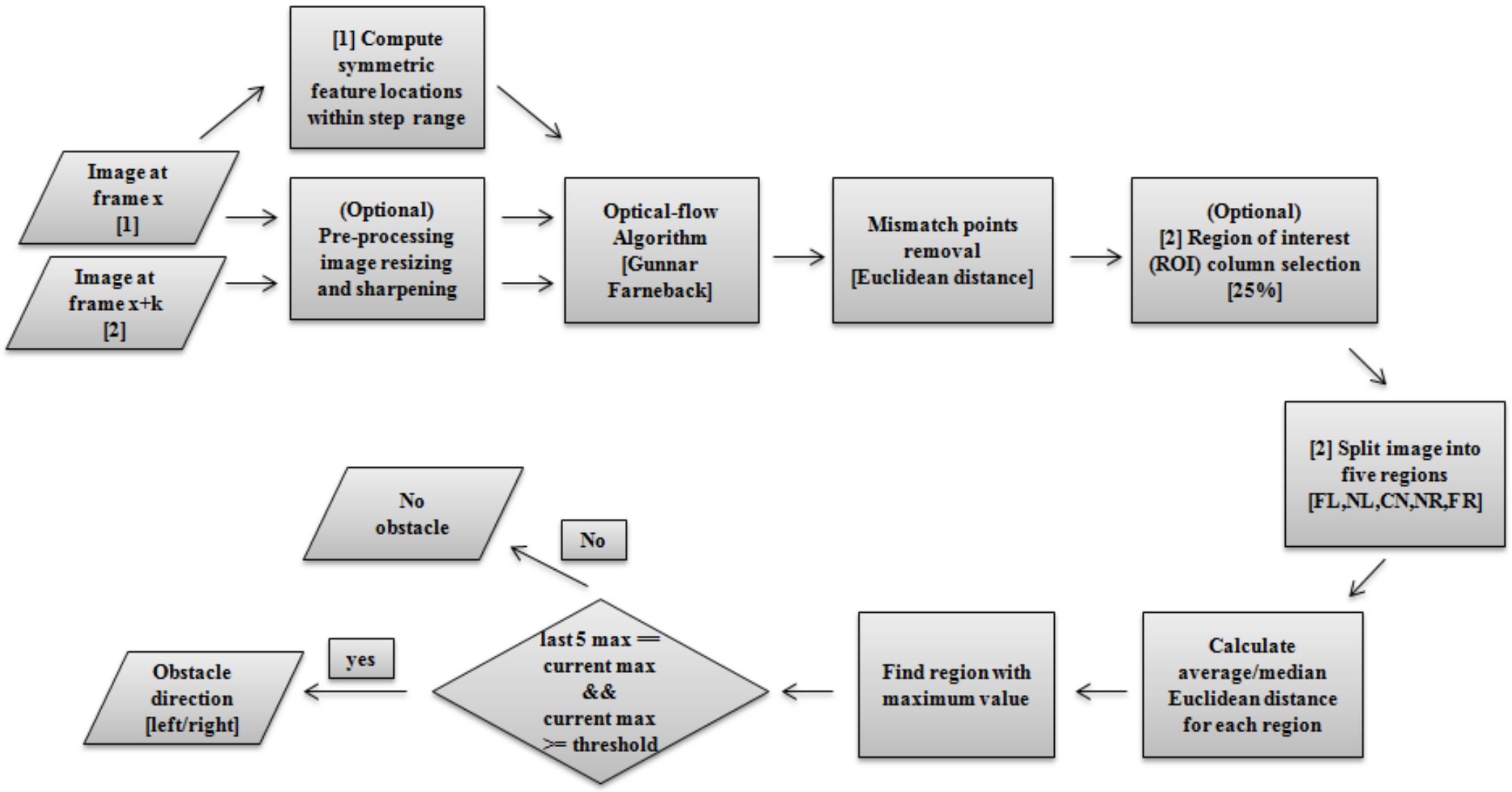

Fig.4 Obstacle detection and avoidance flow chart

Fig.4 shows the flowchart of the proposed obstacle detection algorithm in which its inputs are two consecutive camera frames (in other words, images) and outputs are the existence and direction of obstacle with respect to Ar.drone's view field. The algorithm is based on motion estimation of optical flow with pre/post processing steps for problem utilization and describe in details as following:

I. Set of optional pre-processing steps for both input frames (i.e. decrease image's size for further real-time computation, convert a color space from RGB into HSV, LAB, or normalized RGB for vision-based input enhancement, sharpen the spatial details of input image for getting better image features used in optical flow's point matching operation).
II. Symmetric allocation for point of 1st input frame with respect to step-range parameter.
III. Application of dense-based optical flow algorithm (implemented by Gunnar Farneback ‘03) to compute an optical flow map from two input images and then find the matching points of 2nd image with corresponding previously selected points of 1st image.
IV. Discarding mismatch points based on Euclidean measure in order to have a correct aggregate calculation in next steps of proposed algorithms.
V. Optional selection of regions of intersect on image columns by picking out robot's desired centered field of view (25% starting from image center) in order to have more suitable robot-guidance based on its size.
VI. Splitting the processed image into five equally-sized regions (far left, near left, center, near right, far right) to find obstacle movement's decision based on its position.
VII. Calculation of mean/median magnitude value of all non-discarded points for each region (mean: general averaging representation for region's points, median: alternative representation with discarding outlier points). And then selection of one region among five ones with maximum median value.
VIII. Determination obstacle criterion by checking maximum regions in last 5 iteration and thresholding current maximum value in such that algorithm's output is obstacle position in maximum region if exists. Afterwards, the algorithm output will be used as partial input for obstacle avoidance by moving in the opposite direction of obstacle position (right/left movement if obstacle exists, forward movement if not).

# 6 Obstacle avoidance

The objective of obstacle avoidance is to protect the drone from collision with obstacles in its way when it flies. The obstacle avoidance algorithm used in this project is based on [14] where only the obstacle detected instantly is avoided in reactive obstacle avoidance low. Given that a well-defined obstacle detection algorithm is implemented, the obstacle can be autonomously avoided by sending appropriate reactive control commands for the dynamic model of the Drone. The flow diagram of the overall avoidance is given in Fig 5.

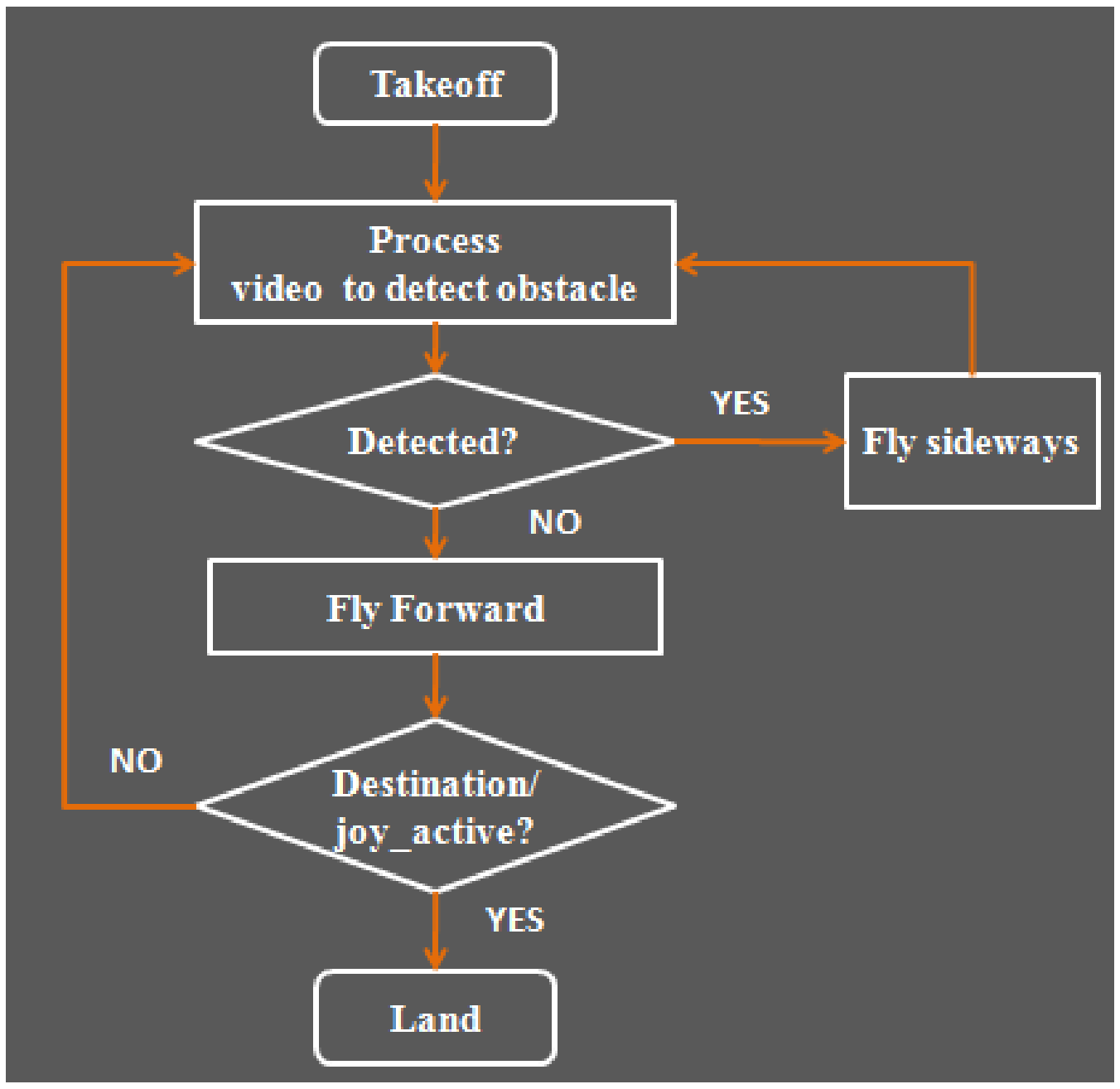

Fig.5 The flow diagram of the overall avoidance

The first step of the flow diagram is the *Takeoff* where the motors of the drone start and the drone takeoff from the ground. After *takeoff*, the video process module will start processing the frames captured by the front camera to detect obstacle. Next to the obstacle detection block, there is a decision box. The decision box will define the next action based on the output of the obstacle detection. If obstacle detected, the drone will fly sideways to avoid the obstacle and return to the obstacle detection state to check if there is obstacle in the front before starting to fly straight forward. But if there is no obstacle detected, the state will shift to *Fly Froward* mode and the Drone start flying forward.

While the drone is flying forward, the Destination/Joy_active state checks if the drone reaches its destination or the Joystick mode is activated. If one of them satisfied, the drone will hover and wait for the next action from the joystick or it will land. But if the two conditions are not satisfied the state of the algorithm will return back to the obstacle detection to and proceed in the same manner as stated above until it reaches the detonation or the joystick mode is activated.

The Joystick mode is introduced for safety purpose. In case in some point if the drone does not react as expected to avoid an obstacle, the joystick override the autonomous flying and let the drone to hove until the next action is given by the joystick. This will help to protect the drone from any damage.

Both the obstacle detection and avoidance algorithms are developed in ROS (Robotic Operating System). ROS is a software framework that enables the development of robotic applications to interact and control robots. In order to use the ROS, it is important to have Linux operating system. Consequently it is important to have a ROS driver that can be used as interface between the drone and the algorithms developed in our computer. The ROS driver for AR.Drone used in this project is the “ardrone_autonomy” developed in [15]. This driver contains all the necessary topics to send and read information from AR.Drone.

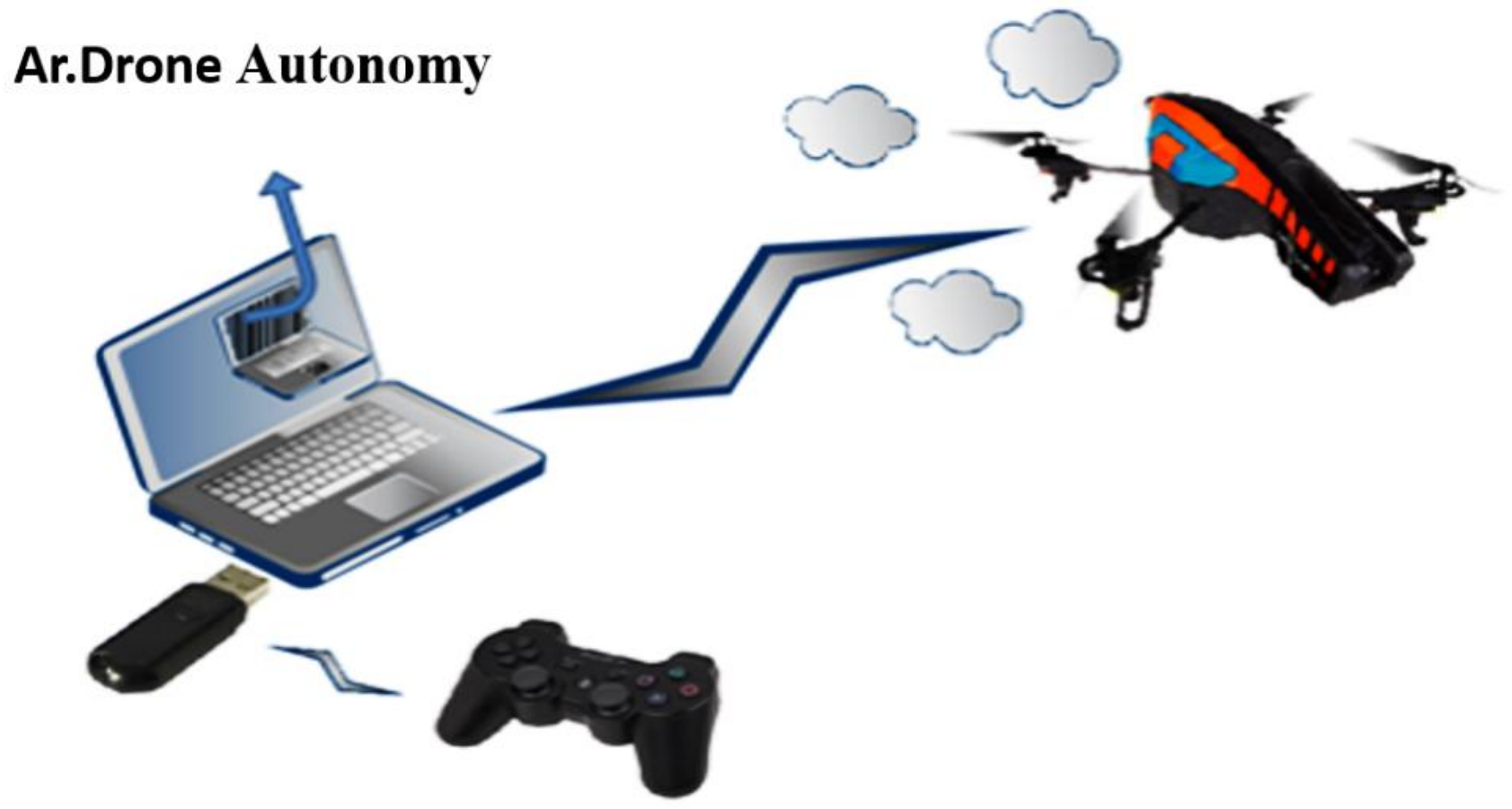

Fig.6 Robot control system

## 6.1 *Reading data from AR.Drone*

The "ardrone_autonomy" driver contains topics to set the data transmission rates between the drone and itself. The AR.Drone transmission frequency is 15Hz. On the other hand the driver can operate in two modes (real-time or fixed rate) to update information. When the real time mode is activated, the driver publishes information instantly. However when the fixed rate mode is activated, the driver catches the most recent received data and publish it at a fixed rate configured by loop rate.

Information received from the AR.Drone will be published to the *ardrone/navdata* topic created by the driver. The *navdata* topic provides the drone status and preprocessed sensory data. The status indicates whether the drone is *flying*, *landing*, *takeoff*, the current type of altitude controller etc. The sensor data contain current yaw, pitch, roll, altitude, battery state and time stamp which is the total time from drone boot-up and many others. Both status and sensory data are updated at 15 Hz rate.

The driver will also create two standard ROS camera interface topics to capture videos from the frontal and bottom cameras of the drone. Before starting to capture video/images using the two cameras, the driver will check if camera calibration information is provided as a set of ROS parameters or appropriate *ardrone_front.yaml* and/or *ardrone_bottom.yaml* to calibrate the cameras. *ardrone_front.yaml* and *ardrone_bottom.yaml* are files containing the calibration of the front and bottom cameras respectively. These two files are included in the "ardrone_autonomy" driver but they should be updated following an instruction given at [16].

## 6.2 Sending commands to AR.Drone

The main objective of the avoidance algorithm is to send control commands to the drone so that the drone will react to avoid the obstacle based on the commands. The ardrone_driver is subscribed to different topics to take values published on these topics and send them to the drone through the Wi-Fi connection. The drone will *takeoff*, *land*, or *emergency stop/reset* by publishing an Empty ROS messages to the following topics:

- ardrone/takeoff
- ardrone/land
- ardrone/reset

The next step after *takeoff* is to fly the drone to a specified goal position or straight forward. This can be easily done by publishing a message of type geometry_msgs::Twist to the cmd_vel topic where the driver is subscribed to it. geometry_msgs::Twist message expresses velocity in free space broken into its linear and angular velocity parts. Both the linear and the angular velocities have three components each representing the linear and angular velocities around x, y and z axis.

As the scope of this project is to fly straight forward and avoid obstacle by flying sideways, the angular velocities are not used as primary controller. The drone can fly forward-backward, left-right, up-down and turn left-right by setting positive and negative velocity values to each component of the linear and angular velocity as shown below.

- +linear.x: fly forward
- -linear.x: fly backward
- +linear.y: fly right
- -linear.y: fly left
- +linear.z: fly up
- -linear.z: fly down
- +angular.z: turn right
- - angular.z: turn left

## 6.3 Autonomous flying controller

To fly the AR.Drone autonomously two packages are developed. The first package contains the algorithms to detect the obstacle while the second package contains all the control algorithms to fly and avoid the obstacle based on the response from the obstacle detection package. The low-level communication hierarchy between the *ardrone_driver* and the implemented packages for obstacle detection and avoidance are given in Fig.7

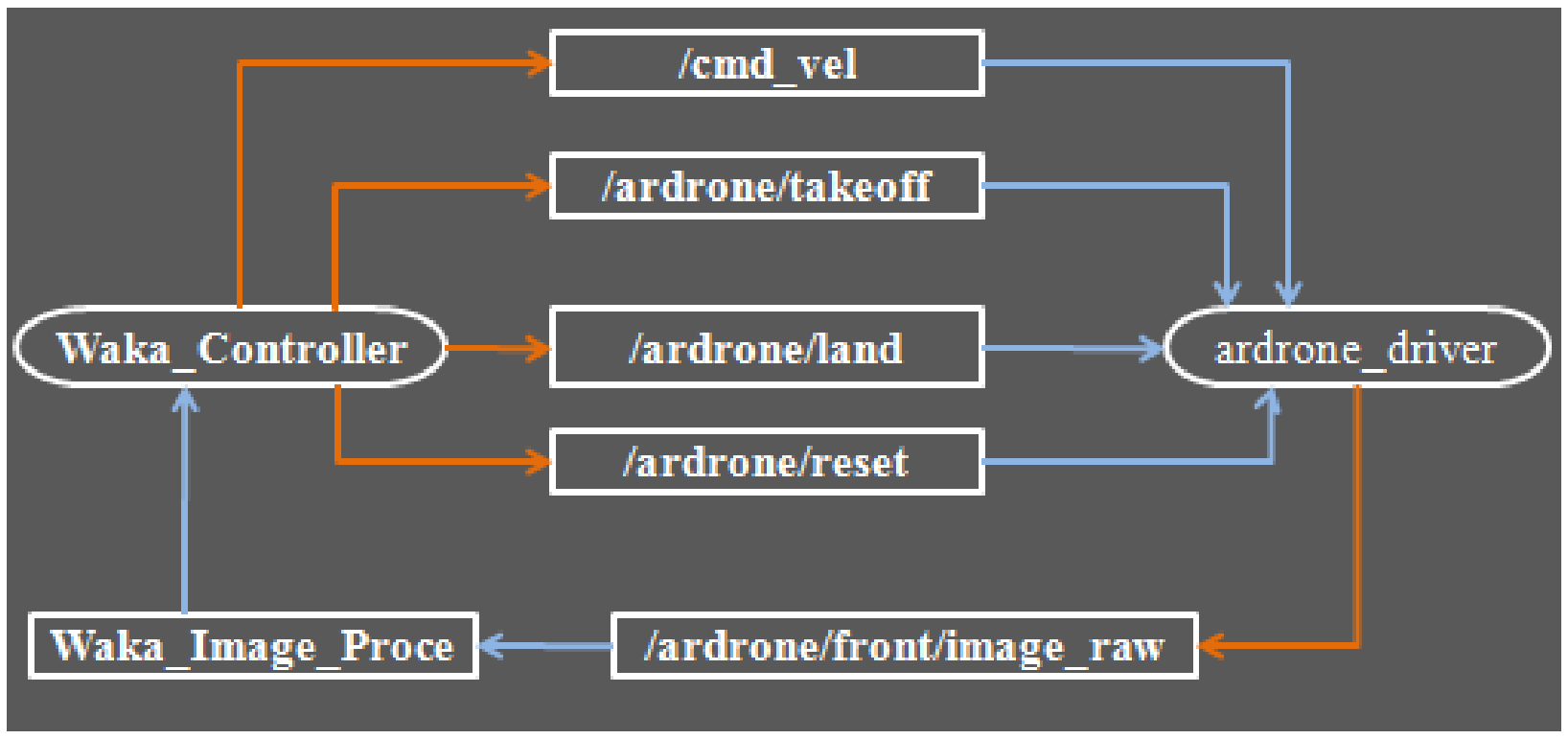

Fig.7 The low-level communication hierarchy between the ardrone_driver and the implemented packages for obstacle detection and avoidance

The hierarchy shows that the *Waka_controller* package publishes the velocity commands to takeoff, land, reset and flies to the given topics. The *ardrone_driver* subscribed to the topics, take the instant velocity commands and transmit them to the drone. On the other hand, the *ardrone_driver* publishes the video frames received from the drone instantly at the specified rate to the *ardrone/front/image_row topic*. The *Waka_Image_Proce* package responsible to analyze and detect the obstacle, takes the video frames and send a signal to the *Waka_controller* package based on the output of the image processing. In this project, the outputs of the *Waka_controller* are defined to be -1, 0 and 1. This will help for the controller to react quickly. -1 and 1 represents an obstacle detected in the left half and right half of the image with respect to the center respectively. Then the controller will avoid the obstacle by flying sideways in opposite to the location of the obstacle. 0 represents no obstacle detects. The image processing package publishes the state at a rate of 15Hz.

The straight forward speed is defined to be 1m/s. This speed will give enough time for the image processing to analyze the image sequences and give the result to the controller to avoid an obstacle. The sideway speed is set to be 1.2 m/s for only 1 sec. This will help the drone to avoid an obstacle/trunk size of 1m. The sideway speed and duration depends on the size of the obstacle/trunk. For example if the maximum size of the obstacle is less than 30 cm, it will be safe even if the sideway fly time is reduced to 0.5 sec. As mentioned above the values of the forward and sideway velocities are assigned to the geometry_msgs::Twist parameters which are *linear.x* and *linear.y* then published to the *cmd_vel* topic. During takeoff, land and reset the Waka_controller publishes an empty ROS message to the corresponding topics. Then *ardrone_driver* send all the velocity commands via the wireless network to the physical drone, which in turn send back a video stream and other navigation data

## *6.4 Joystick integration*

As described in the flowchart of the obstacle avoidance, integrating the joystick with the controller is very important for safety reason to override the autonomously flying mode when the algorithm reacts in unexpected way due to the failure of the obstacle detection. At the same time, the drone can let fly straight forward until any button of the joystick is pressed. The communication hierarchy between each topic after adding the joystick is given in Fig.8.

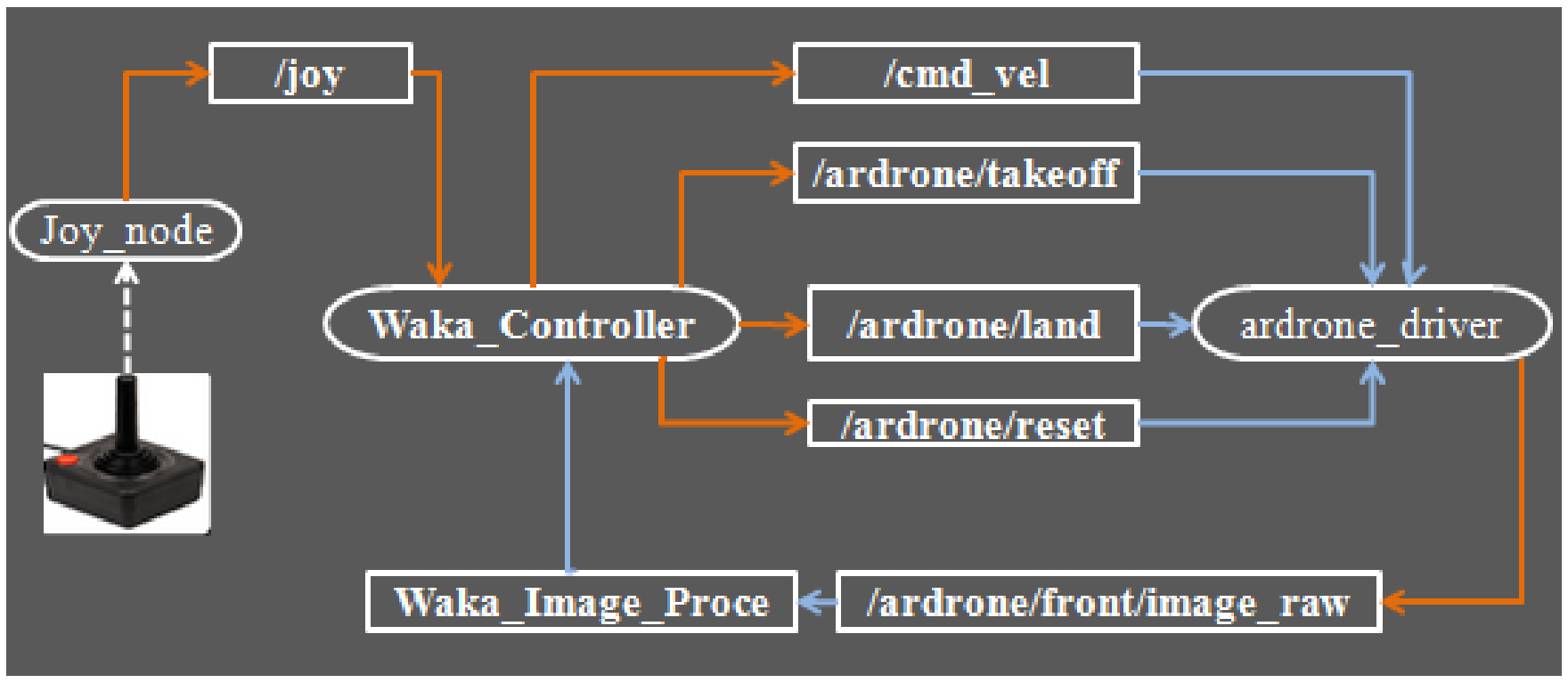

Fig.8 The communication hierarchy between each topic after adding the joystick

In order to integrate the joystick with the controller, it is important to have a joystick driver which is used as interface between the joystick and the ROS. For this project we download a package from [17]. This package contains the joystick driver for interfacing joystick with ROS and converts all joystick events into ROS messages.

# 7 Experiment and Results

## *7.1 Environment and setup*

To train and test the algorithms for obstacle detection and avoidance, indoor and outdoor environments are considered. To train the obstacle detection, indoor and outdoor offline videos are used. These videos are captured using the monocular camera mounted on the drone by flying the drone using our tablets. Different feature extraction methods are used to detect the obstacle.

First, a SURF feature extraction is implemented to extract features for the detection and estimation of the time to collision, but as the result was not good due to insufficient number of features on target objects (tree trunks in outdoor environment or trash bins in indoor environment) which lead to a necessary requirement for putting a high-texture template on target objects.

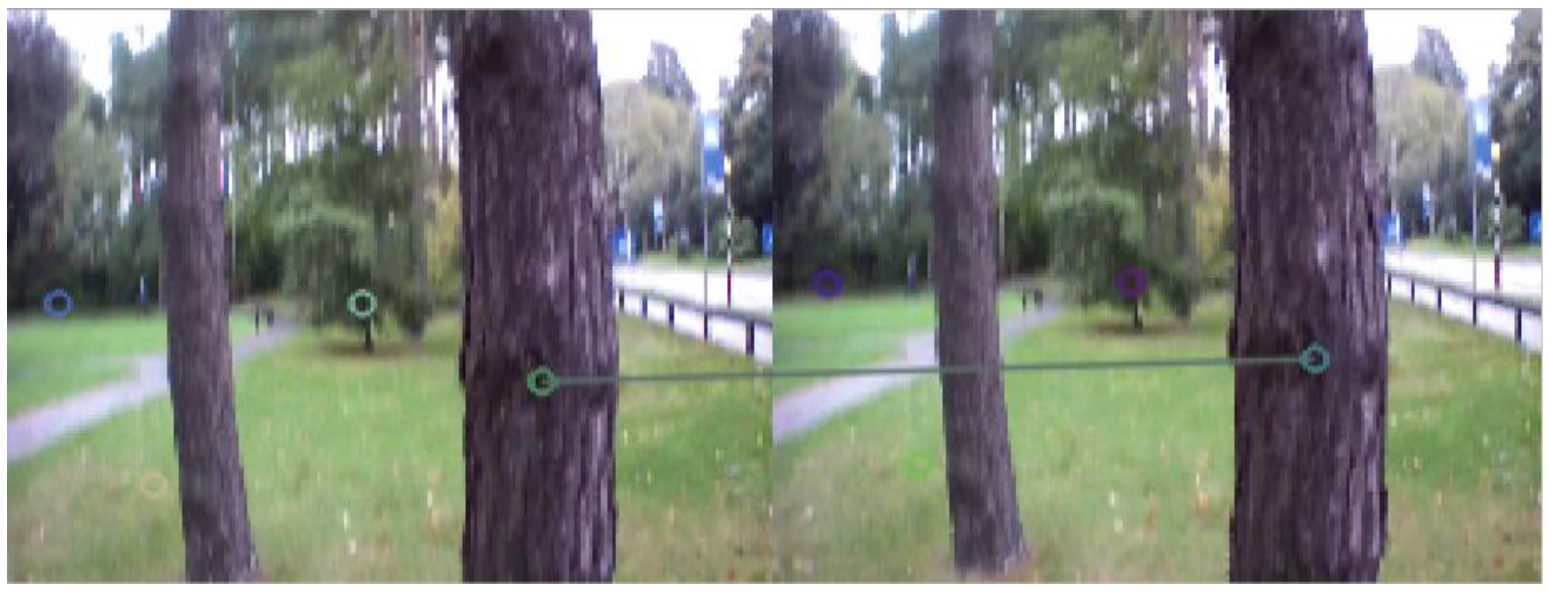

Fig.9 Outdoor result of SURF-based algorithm [9]

We tried to focus on the line detection algorithm so that getting advantages of detection of vertical lines result in finding size expansion of nearby target objects depending on its straight contours, but our experiments as shown in fig.10 have too noisy detected lines and needs specific parameters adjustments to get acceptable lines for target objects.

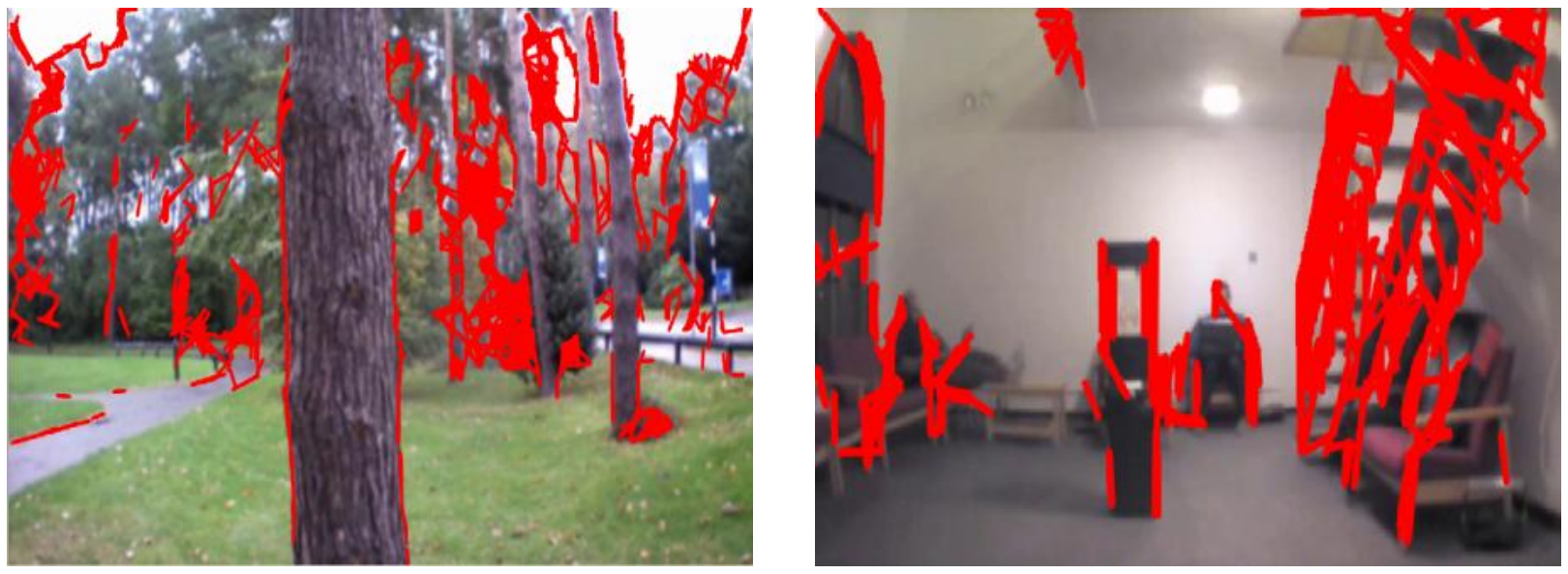

Fig.10 Outdoor and indoor results of line detection algorithm

At the end, we finally decided to use the optical flow algorithm after reviewing some literatures and observe some advantages of its detection. But even though the results of optical flow as shown in fig.11&12 are quite acceptable in motion estimation for semi-dense user-defined features compare with previous ones, although the time-to-collision (TTC) is high-challenged measure in nowadays experiments.

To train the obstacle avoidance controller, a virtual obstacle is assumed in the environment and let the drone to react to the virtual obstacle when it reaches in the position of the estimated virtual obstacle. After successful reaction of the obstacle avoidance for the virtual obstacle, a physical obstacle is used to estimate and train the possible sideway speed and duration. The controller is also trained in outdoor environment to react quickly when obstacle is detected, regardless of the effect of the external windy pressure of the environment. To test the integrated detection and avoidance algorithms, a trash bins are used as an obstacle to detect and avoid it while the drone is flying. High-illuminated room light are used during all training and testing trails.

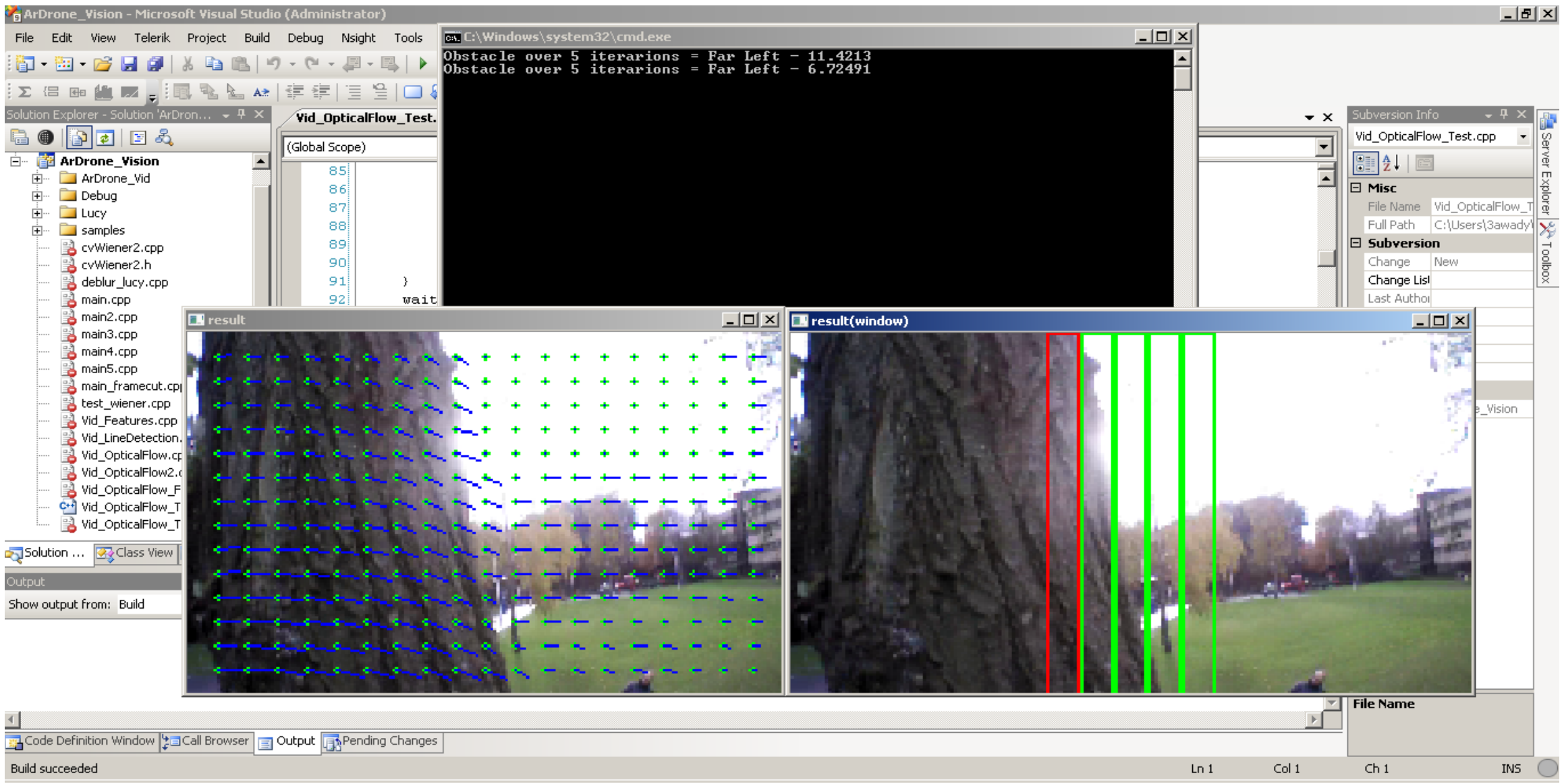

Fig.11 Outdoor result of proposed optical-flow algorithm

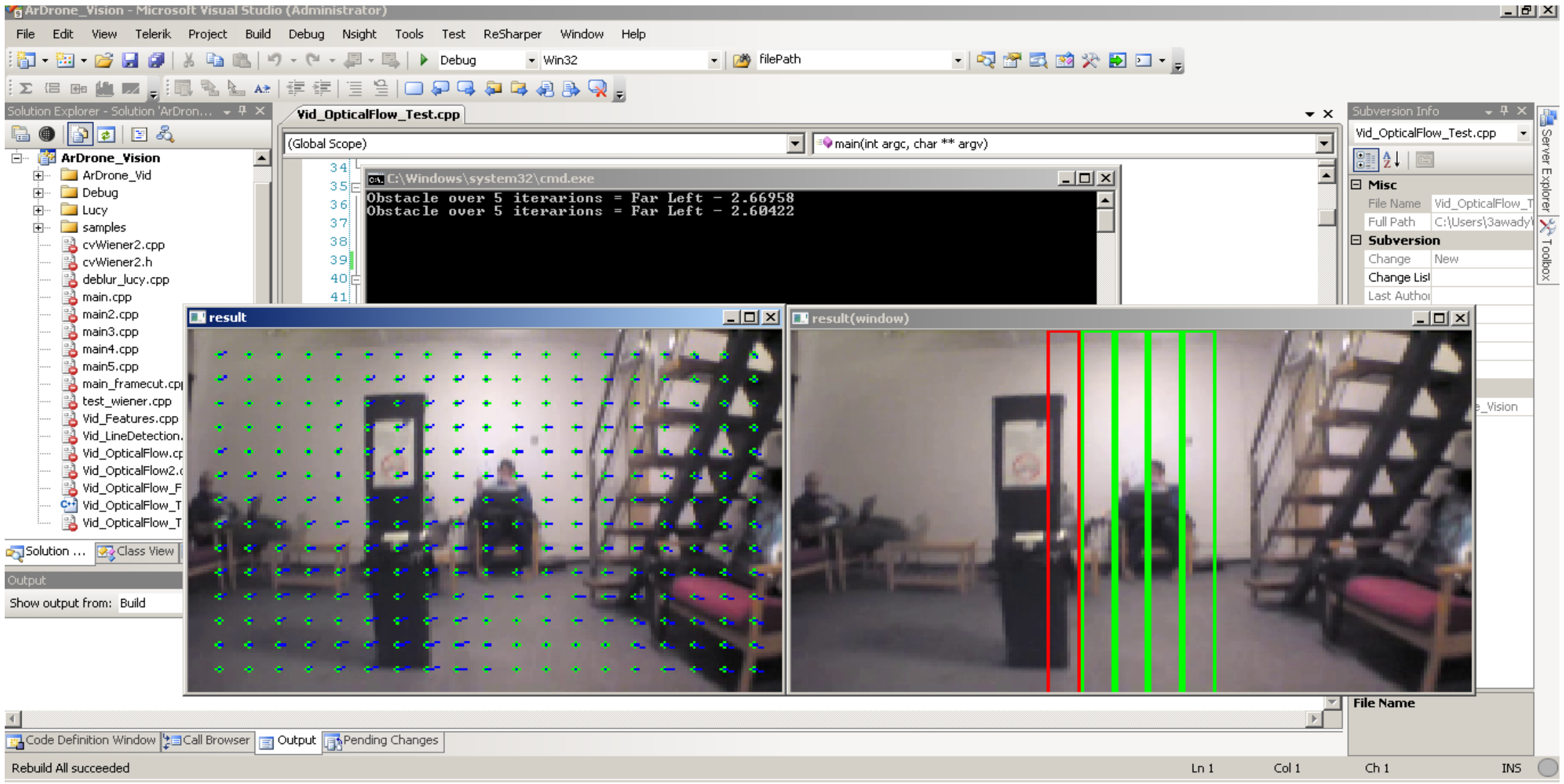

Fig.11 Indoor result of proposed optical-flow algorithm

## 7.2 *Test and results*

The test scheme is divided into two consecutive phases. The first phase includes the individual test of the detection and avoidance for the offline videos and virtual obstacles. The second phase was after combining and integrating the detection and avoidance algorithms by flying in real time. The results of each phase are summarized in the following tables:

| | Indoor | | | Outdoor | | |
|---|---|---|---|---|---|---|
| | Success | Failure | Total | Success | Failure | Total |
| Offline Test | 7 | 5 | 12 | 8 | 4 | 12 |
| Online Test | 7 | 3 | 10 | - | - | - |

The test results show that the performance of the system depends on the performance of the obstacle detection. If the obstacle is not detected and passes the signal to the control algorithm in the right time, the obstacle will not be avoided and the drone will collide with the obstacle. From the observation, the main failure is the estimation of the time to collision, such that putting low-valued algorithm parameters results high-sensitive obstacle detection with any frontal objects. Sometimes, it detects the obstacle from distance and in other times it detects the obstacle when the drone is almost near the obstacle. The first case is not a problem. But in the latter case, if the obstacle is detected when the distance between the obstacle and drone is less than 0.5 meter, the response time will be delayed as the system needs some microseconds to pass the information to the controller then to the *ardrone_driver* and reaches to the drone through the Wi-Fi network. It can be minimized this problem by decreasing the forward speed of the drone to be less than 0.5m/s. But if the forward speed is reduced, the drone angular speed around the y axis will not be stable due to the frontal pressure. This unstable angular velocity introduces a blurry effect to the frames of the video captured.

# 8 Project Management

The project is divided into two modules (detection and avoidance). Each one is separately implemented in offline mode for the detection and using virtual obstacle for the avoidance. Our work plan has been scheduled as following:

| **WORK PLAN** | | | | | | | | |
|---|---|---|---|---|---|---|---|---|
| **Activities** | **September 26,2013 to December 06, 2013** | | | | | | | |
| | **Weeks** | | | | | | | |
| | 3 to 5 | 6 | 7 | 8 | 9 | 10 | 11 | 12 |
| Resource collecting, installation, configuration and analyzing | ■ | | | | | | | |
| Detection Implementation using offline videos | | ■ | ■ | ■ | ■ | | | |
| Avoidance Implementation using virtual obstacle | | ■ | ■ | ■ | ■ | | | |
| Integrating both detection and avoidance, Testing | | | | | | ■ | ■ | ■ |
| Report writing | | | | | | ■ | ■ | ■ |
| Presentation | | | | | | | ■ | |
| Demonstration | | | | | | | | ■ |

# 9 Conclusion

Obstacle avoidance and detection using monocular camera is challenging task, since obstacles cannot be directly detected. The drone has several capabilities to enable automatic obstacle detection such as a high resolution frontal camera, a bottom camera to provide its speed with respect to the ground. However, after some trails we have seen that the drone is really affected by ambient environments i.e. air currents which makes the drone slightly unstable, also lower velocity cause unstable behavior during flying that's why we keep the velocity as 1 m/s. an important feature of the drone is high resistance against any un expected crashing. The obstacle detection is a bit difficult because of non-smooth movement of the drone resulting average video capturing. Another problem with the drone is error accumulation such as after some operations we have to reset it otherwise unwanted behavior will occur. In proposed algorithm, a semi-dense optical flow is used in order to detect frontal obstacles, the algorithm can detect those obstacles but due to lighting variation conditions and in-accurate calculation of the time to collision, the algorithm results are not always the same for the same object across consecutive iterations. The algorithm efficiency is tested offline in indoor and outdoor environments while outdoor results are better than indoor results. That makes sense as there is more enough texture information on the trees trunks than indoor objects.

# 10 Future works and recommendations

Drone's frontal camera is not sufficient enough finding how far obstacle is (time-aspect or distance-aspect) using optical-flow algorithm, such that integrating different external sensors (laser-range sensor or stereo-vision camera) with drone robot and combining applications of different detectors (optical-flow, SURF, etc...) will improve the proposed algorithm. In obstacle detection, we recommend two implementation improvements: (1) deciding the drone's movement in specific frame using optical flow maps of previous frames. (2) instead of neglecting deblurring frames, applying a deblurring algorithm (Richardson-Lucy algorithm [18] or Wiener filter [19]) by finding suitable motion kernel based drone's behavior. In obstacle avoidance, we suggest to make drone following m-line to finite goal depending on well-known path planning algorithm.